%% file: main.tex
\def\year{2021}\relax
\documentclass[letterpaper]{article} 
\usepackage{aaai21}  
\usepackage{times}  
\usepackage{helvet} 
\usepackage{courier}  
\usepackage[hyphens]{url}  
\usepackage{graphicx} 
\usepackage{natbib}  
\usepackage{caption} 
\usepackage{amsmath,amsthm,amssymb}
\usepackage{booktabs}
\usepackage[capitalize]{cleveref}

\DeclareMathOperator{\Bernoulli}{Bernoulli}

\newcommand{\xb}{\mathbf{x}}
\newcommand{\zb}{\mathbf{z}}
\newcommand{\given}{\,|\,}
\newcommand{\ptheta}{p_{\theta}}
\newcommand{\qphi}{q_{\phi}}
\newcommand{\E}{\mathbb{E}}
\newcommand{\kl}{D_\mathrm{KL}}

\title{Planning from Pixels in Atari with Learned Symbolic Representations}
\newcommand*\samethanks[1][\value{footnote}]{\footnotemark[#1]}
\author {
        Andrea Dittadi,\thanks{Equal contribution.}\ 
        Frederik K. Drachmann,\samethanks\ 
        Thomas Bolander
        \\
}
\affiliations {
    Technical University of Denmark,
    Copenhagen, Denmark \\
    adit@dtu.dk, fdrachmann@hotmail.dk, tobo@dtu.dk
}

\usepackage[switch]{lineno}

\begin{document}

\maketitle

\begin{abstract}
Width-based planning methods have been shown to yield state-of-the-art performance in the Atari 2600 domain using pixel input. One successful approach, RolloutIW, represents states with the B-PROST boolean feature set. An augmented version of RolloutIW, $\pi$-IW, shows that learned features can be competitive with handcrafted ones for width-based search. 
In this paper, we leverage variational autoencoders (VAEs) to learn features directly from pixels in a principled manner, and without supervision. 
The inference model of the trained VAEs extracts boolean features from pixels, and RolloutIW plans with these features. The resulting combination outperforms the original RolloutIW and human professional play on Atari 2600 and drastically reduces the size of the feature set.
\end{abstract}

\input{sections/introduction}

\input{sections/background}

\input{sections/VAE-IW}

\input{sections/experiments}

\input{sections/related_work}

\input{sections/conclusion}

\bibliography{main.bib}

\clearpage

\appendix

\input{sections/appendix}

\end{document}

%% file: sections/introduction.tex
\section{Introduction}
Width-based search algorithms have in the last few years become among the state-of-the-art approaches to  automated planning, e.g.\ the original Iterated Width (IW) algorithm~\cite{iworginal}. As in propositional STRIPS planning, states are represented by a set of propositional literals, also called boolean \emph{features}. The state space is searched with breadth-first search (BFS), but the state space explosion problem is handled by pruning states based on their \emph{novelty}. First, a parameter $k$ is chosen, called the \emph{width parameter} of the search. Searching with width parameter $k$ essentially means that we only consider $k$ literals/features at a time. A state $s$ generated during the search is called \emph{novel} if there exists a set of $k$ literals/features not made true in any earlier generated state. Unless a state is novel, it is immediately pruned from the search. Clearly, we then reduce the size of the searched state space to be exponential in $k$. It has been shown that many classical planning problems, e.g.\ problems from the International Planning Competition (IPC) domains, can be solved efficiently using width-based search with very low values of $k$.  

The essential benefit of using width-based algorithms is the ability to perform semi-structured (based on feature structures) exploration of the state space, and reach deep states that may be important for achieving the planning goals. In classical planning, width-based search has been integrated with heuristic search methods, leading to Best-First Width Search~\cite{BFSW} that performed well at the 2017 IPC. Width-based search has also been adapted to reward-driven problems where the algorithm uses a simulator for interacting with the environment \cite{ijcai2017-600}. This has enabled the use of width-based search in reinforcement learning environments such as the Atari 2600 video game suite, through the Arcade Learning Environment (ALE)~\cite{DBLP:journals/corr/abs-1207-4708}. There have been several implementations of width-based search directly using the RAM states of the Atari computer as features \cite{ram_model,ram_IW_shleyfman,dominatedSequnce}.

Motivated by the fact that humans do not have access to RAM states when playing video games, methods for using these algorithms from screen pixels have been developed. \citet{rolloutiw-orginal} propose a modified version of IW, called RolloutIW, that uses pixel-based features and achieves results comparable to learning methods in almost real-time on ALE. The combination of reinforcement learning and RolloutIW in the pixel domain is explored by \citet{RLIW}, who propose to train a policy to guide the action selection in rollouts. 

Width-based algorithms are highly exploratory and, not surprisingly, significantly dependent on the quality of the features defined. 
\citet{rolloutiw-orginal} use
a set of low-level, color-based features called B-PROST~\cite{liang},
designed by humans to achieve good performance in ALE. Integrating feature learning
into the algorithms themselves rather than using hand-crafted features is an interesting challenge---and a challenge that will bring the algorithms more on a par with humans learning to play those video games. The features used by \citet{RLIW} were the output features of the last layer of the policy network, improving the performance of IW. With the attempt to learn efficient features for width-based search, and inspired by the results of \citet{RLIW}, this paper investigates the possibility of a more structured approach to generate features for width-based search using deep generative models. 

More specifically, in this paper we use variational autoencoders (VAE)---latent variable models that have been widely used for representation learning \cite{kingma2019introduction} as they allow for approximate inference of the latent variables underlying the data---to learn propositional symbols directly from pixels and without any supervision.
We then investigate whether the learned symbolic representation can be successfully used for planning in the Atari 2600 domain. 

We compare the planning performance of RolloutIW using our learned representations with RolloutIW using the handcrafted B-PROST features and report human scores as a baseline.
Our results show that our learned representations lead to better performance than B-PROST on RolloutIW (and often outperform human players), although the B-PROST features also contain temporal information (how the screen changes), whereas we only extract static features from individual frames. 
Apart from improving scores in the Atari 2600 domain, we also significantly reduce the number of features (by a factor of more than $10^3$). 
We also investigate in detail (with ablation studies) which factors seem to lead to good performance on games in the Atari domain.

The main contributions of our paper are hence:
\begin{itemize}
    \item We propose a novel combination of variational autoencoders and width-based planning algorithms.
    \item We train VAEs to learn propositional symbols for planning, from pixels and without any supervision.
    \item We run a large-scale evaluation on Atari 2600 where we compare the performance of RolloutIW with our learned representations to RolloutIW with B-PROST features.
    \item We investigate with ablation studies the effect of various hyperparameters in both learning and planning.
    \item We show that our learned representations lead to higher scores than using B-PROST, even though (i) our features don't have any temporal information and (ii) our models are trained from data collected by RolloutIW with B-PROST, thus limiting the richness of the dataset.
\end{itemize}
Below, we provide the required background on RolloutIW, B-PROST, and VAEs, present the original approach of this paper, discuss our experimental results, and conclude.

%% file: sections/background.tex
\section{Background}

\paragraph{Iterated width.}
Iterated Width (IW)~\cite{iworginal} is a blind-search planning algorithm in which the states are represented by sets of boolean features (sets of propositional atoms). The set of all features/atoms is the \emph{feature set}, denoted $F$. IW is an algorithm parameterized by a \emph{width parameter} $k$. We use IW($k$) to denote IW with width parameter $k$. 
Given an initial state $s_0$, IW($k$) is similar to a standard breadth-first search (BFS) from $s_0$ except that, when a new state $s$ is generated, it is immediately pruned if it is not novel. A state $s$ generated during search is defined to be \emph{novel} if there is a $k$-tuple of features (atoms) $t = (f_1, \ldots, f_k) \in F^k$ such that $s$ is the first generated state making all of the $f_i$ true ($s$ making $f_i$ true of course simply means $f_i \in s$, and the intuition is that $s$ then ``has'' the feature $f_i$).  In particular, in IW($1$), the only states that are not pruned are those that contain a feature $f \in F$ for the first time during the search.
The maximum number of generated states is exponential in the width parameter (IW($k$) generates $O(|F|^k)$ states), whereas in BFS it is exponential in the number of features/atoms.

\paragraph{Rollout IW.}
RolloutIW($k$)~\cite{rolloutiw-orginal} is a variant of IW($k$) that searches via rollouts instead of doing a breadth-first search, hence making it more akin to a depth-first search. A \emph{rollout} is a state-action trajectory $(s_0, a_0, s_1, a_1, \ldots)$ from the initial state $s_0$ (a path in the state space), where actions $a_i$ are picked at random. It continues until reaching a state that is not novel. A state $s$ is \emph{novel} if it satisfies one of the following: 1) it is distinct from all previously generated states and there exists a $k$-tuple of features $(f_1,\dots,f_k)$ that are true in $s$, but not in any other state of the same depth of the search tree or lower; or 2) it is an already earlier generated state and there exists a $k$-tuple of features $(f_1,\dots,f_k)$ that are true in $s$, but not in any other state of lower depth in the search tree. The intuition behind case 1 is that the new state $s$ comes with a combination of $k$ features occurring at a lower depth than earlier encountered, which makes it relevant to explore further. In case 2, $s$ is an existing state containing the lowest occurrence of one of the combinations of $k$ features, again making it relevant to explore further. When a rollout reaches a state that is not novel, the rollout is terminated, and a new rollout is started from the initial state. The process continues until all possible states are explored, or a given time limit is reached. After the time limit has been reached, an action with maximal expected reward is chosen (the method was developed in the context of the Atari 2600 domain, making use of a simulator to compute action outcomes and rewards). The chosen action is executed, and the algorithm is repeated with the new state as the initial state. 

RolloutIW is an anytime algorithm that will return an action independently of the time budget. In principle, IW could also be used as an anytime algorithm in the context of the Atari 2600 domain, but it will be severely limited by the breadth-first search strategy that will in practice prevent it from reaching nodes of sufficient depth in the state space, and hence prevent it from discovering rewards that only occur later in the game~\cite{rolloutiw-orginal}.

\paragraph{Planning with pixels.}
The Arcade Learning Environment (ALE) \cite{DBLP:journals/corr/abs-1207-4708} provides an interface to Atari 2600 video games, and has been widely used in recent years as a benchmark for reinforcement learning and planning algorithms. In the visual setting, the sensory input consists of a pixel array of size $210\!\times\! 160$, where each pixel can have 128 distinct colors. Although in principle the set of $210\! \times\! 160\! \times\! 128$ booleans could be used as features, we will follow \citet{rolloutiw-orginal} and focus on features that capture meaningful structures from the image.

An example of a visual feature set that has proven successful is \emph{B-PROST}~\cite{liang}, which consists of \emph{Basic}, \emph{B-PROS}, and \emph{B-PROT} features. 
The screen is split into $14 \times 16$ disjoint tiles of size $15 \times 10$. 
For each tile $(i,j)$ and color $c$, the basic feature $f_{i,j,c}$ is 1 iff $c$ is present in $(i,j)$.
A B-PROS feature $f_{i,j,c,c'}$ is 1 iff color $c$ is present in tile $t$, color $c'$ in tile $t'$, and the relative offsets between the tiles are $i$ and $j$.
Similarly, a B-PROT feature $f_{i,j,c,c'}$ is 1 iff color $c$ is present in tile $t$ \emph{in the previous decision point}, color $c'$ is present in tile $t'$ in the current one, and the relative offsets between the tiles are $i$ and $j$.
The number of features in B-PROST is the sum of the number of features in these 3 sets, in total 20,598,848.

\paragraph{Variational autoencoders.}
\emph{Latent variable models} (LVMs) are probabilistic models with unobserved variables. The marginal distribution over one observed datapoint $\xb$ is
$\ptheta (\xb) = \int_{\zb} \ptheta (\xb, \zb) d \zb,$
with $\zb$ the unobserved \emph{latent variables} and $\theta$ the model parameters. This quantity is typically referred to as \emph{marginal likelihood} or \emph{model evidence}. A simple and rather common structure for LVMs is:
%
%
\begin{equation*}
    \ptheta (\xb, \zb) = \ptheta (\xb \given \zb) \ptheta (\zb).
\end{equation*}
%
If the model's distributions are parameterized by neural networks, the marginal likelihood is typically intractable for lack of an analytical solution or a practical estimator.

\emph{Variational inference} (VI) is a common approach to approximating this intractable posterior, where we define a distribution $\qphi (\zb \given \xb)$ with \emph{variational parameters} $\phi$.
In the LVM above, for any choice of $\qphi$ we have:
\begin{align*}
    \log \ptheta(\xb) 
    &= \log \E_{\qphi(\zb \given \xb)} \left[ \frac{\ptheta(\xb \given \zb)\ptheta(\zb)}{\qphi(\zb \given \xb)} \right] \\
    & \geq \E_{\qphi(\zb \given \xb)}  \left[ \log \frac{\ptheta(\xb \given \zb)\ptheta(\zb)}{\qphi(\zb \given \xb)} \right] 
    = \mathcal{L}_{\theta,\phi} (\xb)
\end{align*}
where $\mathcal{L}_{\theta,\phi} (\xb)$ is a lower bound on the marginal log likelihood also known as \emph{Evidence Lower BOund} (ELBO).

In contrast to traditional VI methods, where per-datapoint variational parameters are optimized separately, \emph{amortized variational inference} utilizes function approximators like neural networks to share variational parameters across data\-points and improve learning efficiency. In this setting, $\qphi (\zb \given \xb)$ is typically called an \emph{inference model} or \emph{encoder}.
Variational Autoencoders (VAEs)~\cite{kingma2013auto,rezende2014stochastic} are a framework for amortized VI, in which the ELBO is maximized by jointly optimizing the inference model and the LVM (i.e., $\phi$ and $\theta$, respectively) with stochastic gradient ascent.

\paragraph{VAE optimization.}
The ELBO, the objective function to be maximized, can be decomposed as follows:
\begin{align}
    \mathcal{L}_{\theta,\phi} (\xb) 
    &= \E_{\qphi}  \left[ \log \ptheta(\xb \given \zb) \right] - \E_{\qphi}  \left[ \log \frac{\qphi(\zb \given \xb)}{\ptheta(\zb)} \right] \label{eq:elbo} \\
    &= \E_{\qphi}  \left[ \log \ptheta(\xb \given \zb) \right] - \kl(\qphi(\zb \given \xb) \,||\, \ptheta(\zb)) \nonumber
\end{align}
where the first term can be interpreted as negative expected \emph{reconstruction error}, and the second term is the KL divergence from the prior $\ptheta(\zb)$ to the approximate posterior.

When optimizing VAEs, the gradients of some terms of the objective function cannot be backpropagated through the sampling step.
However, for a rather wide class of probability distributions \cite{kingma2013auto}, a random variable following such a distribution can be expressed as a differentiable, deterministic transformation of an auxiliary variable with independent marginal distribution. 
For example, if $z$ is a sample from a Gaussian random variable with mean $\mu_\phi(x)$ and standard deviation $\sigma_\phi(x)$, then $z = \sigma_\phi(x)\, \epsilon + \mu_\phi(x)$, where $\epsilon \sim \mathcal{N}(0,1)$. Thanks to this \emph{reparameterization}, $z$ can be differentiated with respect to $\phi$ by standard backpropagation.
This widely used approach, called \emph{pathwise gradient estimator}, typically exhibits lower variance than the alternatives.

From an information theory perspective, optimizing the variational lower bound~\eqref{eq:elbo} involves a tradeoff between rate and distortion \cite{alemi2017fixing}.
A straightforward way to control the rate--distortion tradeoff is to use the $\beta$-VAE framework \cite{higgins2017beta}, in which the training objective~\eqref{eq:elbo} is modified by scaling the KL term:
%
\begin{align}
    \mathcal{L}_{\theta,\phi, \beta} (\xb) &= \E_{\qphi}  \left[ \log \ptheta(\xb \given \zb) \right] \label{eq:beta_objective_function}\\
    &\qquad - \beta \kl(\qphi(\zb \given \xb) \,||\, \ptheta(\zb)). \nonumber
\end{align}

\paragraph{VAEs with discrete variables.}
Since the categorical distribution is not reparameterizable, training VAEs with categorical latent variables is generally impractical. A solution to this problem is to replace samples from a categorical distribution with samples from a Gumbel-Softmax distribution \cite{jang2016categorical,maddison2016concrete}, which can be smoothly annealed into a categorical distribution by making the temperature parameter $\tau$ tend to~0. Because Gumbel-Softmax is reparameterizable, the pathwise gradient estimator can be used to get a low-variance---although in this case biased---estimate of the gradient.
In this work, we use Bernoulli latent variables (categorical with 2 classes) and their Gumbel-Softmax relaxations.

%% file: sections/VAE-IW.tex
\section{VAE-IW}

Although width-based planning has been shown to be generally very effective for classical planning domains \cite{iworginal,BFSW,ijcai2017-600}, its performance in practice significantly depends on the quality of the features used. Features that better capture meaningful structure in the environment typically translate to better planning results \cite{RLIW}. 
Here we propose VAE-IW, in which representations extracted by a VAE are used as features for RolloutIW. 
The main advantages are:
\begin{itemize}
    \item Significantly fewer features than B-PROST, leading to faster planning and a smaller memory footprint.
    \item Autoencoders, in particular VAEs, are a natural approach for learning meaningful, compact representations from data \cite{bengio2013representation,tschannen2018recent,kingma2019introduction}.
    \item No additional preprocessing is needed, such as background masking~\cite{rolloutiw-orginal}.
\end{itemize}
On the planning side, we follow \citet{RLIW} and use RolloutIW($k$) with width $k=1$ to keep planning efficient even with a small time budget.
In the remainder of this section, we detail the two main components of the proposed method: feature learning and planning.

\paragraph{Unsupervised feature learning.}

For feature learning we use a variational autoencoder with a discrete latent space:
\begin{equation}
    \ptheta(\xb) = \sum\nolimits_{\zb} \ptheta(\xb, \zb) = \sum\nolimits_{\zb} \ptheta(\xb \given \zb) p(\zb)
\label{eq:vaeiw-marginal}
\end{equation}
where the prior is a product of independent Bernoulli distributions with a fixed parameter:
    $p(\zb) = \prod_{i=1}^K p(z_i)$, where $ p(z_i) = \Bernoulli(\mu)$.
Given an image $\xb$, we would like to retrieve the binary latent factors $\zb$ that generated it, and use them as a propositional representation for planning. Since $\ptheta(\xb \given \zb)$ is parameterized by a neural network, and therefore highly nonlinear with respect to the latent variables, inference of the latent variables is intractable (it would require the computation of the sum in~\eqref{eq:vaeiw-marginal}, which has a number of terms exponential in the number of latent variables).

We define an inference model:
\begin{align*}
    \qphi(\zb \given \xb) 
    &= \prod_{i=1}^K \qphi(z_i \given \xb) 
    = \prod_{i=1}^K \Bernoulli((\mu_\phi(\xb))_i)
\end{align*}
to approximate the true posterior $\ptheta(\zb \given \xb)$.
The \emph{encoder} $\mu_\phi$ is a deep neural network with parameters $\phi$ that outputs the approximate posterior probabilities of all latent variables given an image $\xb$.
Using stochastic amortized variational inference, we train the inference and generative models end-to-end by maximizing the ELBO with respect to the parameters of both models.
We approximate the discrete Bernoulli distribution of $\zb$ with the Gumbel-Softmax relaxation~\cite{jang2016categorical,maddison2016concrete}, which is reparameterizable. This allows us to estimate the gradients of the inference network with the pathwise gradient estimator.
Alternatively, the approximate posterior could be optimized directly using a score-function estimator~\cite{williams1992simple} with appropriate variance reduction techniques or alternative variational objectives~\cite{Bornschein2014-my,Mnih2016-vr,Le2018-kj,Masrani2019-nq,lievin2020optimal}.

Note that we are not interested in the best generative model---e.g. in terms of marginal log likelihood of the data or visual quality of the generated samples---as much as in representations that are useful for planning.
Thus, ideally, we would directly optimize $\beta$ for planning performance. In practice, however, we assume that a necessary (but not sufficient) condition for a representation to be useful in our case is that it contains enough information about the entities on the screen.
Following \citet{burgess2018understanding}, we control the tradeoff between the competing terms in~\eqref{eq:elbo} by using~\eqref{eq:beta_objective_function} and decreasing $\beta$ until good quality reconstructions are obtained.

\paragraph{Planning with learned features.}

After training, the inference model $\qphi(\zb \given \xb)$ yields approximate posterior distributions of the latent variables. The binary features for downstream planning can be obtained by sampling from these distributions or by deterministically thresholding their probabilities given a fixed threshold $\lambda \in (0, 1)$.
In this work we choose the latter approach as it empirically yields more stable features for planning.
Overall, this provides a way of efficiently computing a compact, binary representation of an image, which can in turn be interpreted as a set of propositional features to be used in symbolic reasoning systems like IW or RolloutIW. 

The planning phase in VAE-IW is based on RolloutIW, but uses the features extracted by the inference network instead of the B-PROST features. Note that while the B-PROST features include temporal information, the VAE features are computed independently for each frame. This should, everything else being equal, give an advantage to planners that rely on B-PROST features.

%% file: sections/experiments.tex
\section{Experiments}

We evaluated the proposed approach on a suite of 55 Atari 2600 games.\footnote{
Code available at \url{https://github.com/fred5577/VAE-IW}}
We separately trained a VAE for each domain by maximizing~\cref{eq:beta_objective_function} using stochastic gradient ascent.
To train the VAEs, we collected 15,000 frames by running RolloutIW(1) with B-PROST features, and split them into a training and validation set of 14,250 and 750 images.

The encoder consists of 3 convolutional layers interleaved with residual blocks. The single convolutional layers downsample the feature map by a factor of 2, and each residual block includes 2 convolutional layers, resulting in 7 convolutional layers in total. The decoder mirrors such architecture, and uses transposed convolutions for upsampling. Further architectural details are provided in the Appendix.
The inference network outputs a feature map of shape $H \times W \times C$ which represents the approximate posterior probabilities of the latent variables. Thus, unlike traditional VAEs where latent variables do not carry any spatial meaning, in our case they are spatially arranged~\cite{vahdat2020nvae}.

As outlined in the previous section, the Bernoulli probabilities computed by the encoder are thresholded, and the resulting binary features are used as propositional representations for planning in RolloutIW(1), instead of the B-PROST features.
For the planning phase of VAE-IW(1), we use RolloutIW(1) with partial caching and risk aversion (RA) as described by \citet{rolloutiw-orginal}. With partial caching, after each iteration of RolloutIW(1), the branch of the chosen action is kept in memory. Risk aversion is attained by multiplying all negative rewards by a factor $\alpha \gg 1$ when they are propagated up the tree.

Because of their diversity, Atari games vary widely in complexity.
We empirically chose a set of hyperparameters that performed reasonably well on a small subset of the games, assuming this would generalize well enough to other games.
Following previous work~\cite{iworginal,rolloutiw-orginal}, we used a frame skip of 15 and a planning budget of 0.5s per time step.
We set an additional limit of 15,000 executed actions for each run, to prevent runs from lasting too long (note that this constraint is only applied to our method).
We expect that better results can be obtained by performing an extensive hyperparameter search on the whole Atari suite.
See the Appendix for further implementation details.

\subsection{Main Results}

\begin{table}[!t]
    \centering
    \begin{minipage}{\columnwidth}
    \input{tables/human_comp}
    \end{minipage}
\end{table}

\begin{figure*}[!ht]
    \centering
    \includegraphics[width=\textwidth]{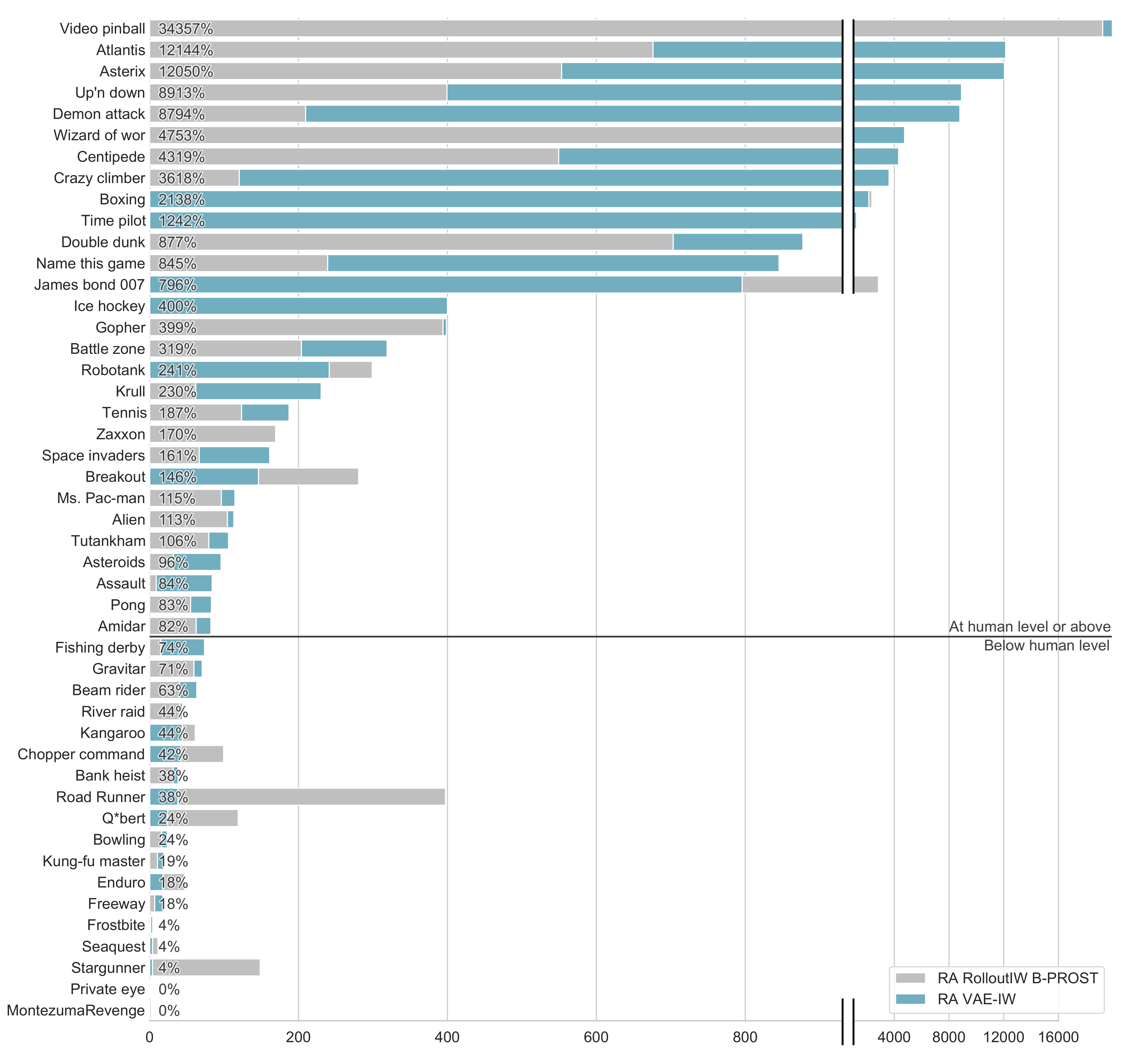}
    \caption{Comparison of the risk-averse variants of VAE-IW(1) (ours, using {VAE} features) and RolloutIW(1) B-PROST (using {B-PROST} features). Following \citet{minh}, the performance of both methods is normalized with respect to a professional human game tester (100\% level) and random play (0\%) as: $100 \times (
    \text{VAE}-\text{random play})/(\text{human score}-\text{random play})$. RA VAE-IW(1) obtains the highest score among width-based approaches in most games, and it performs at a level that is superior to or comparable with professional human play. The reported percentages are for VAE features.}
    \label{fig:cool_plot}
\end{figure*}

In Table~\ref{tab:atari-lookahead-0.5}, we compare the planning performance of our RA VAE-IW(1) to IW(1) with B-PROST features, RA RolloutIW(1) with B-PROST features, and human play \citep{liang}. 
In Figure~\ref{fig:cool_plot}, we normalize the scores of width-based planning methods using scores from random and human play, as in \citet{rolloutiw-orginal}. Note that human play is only meant as a baseline: since humans do not have access to a simulator, a direct comparison cannot be made. 
Following the literature, we report the average score over 10 runs, with each run ending when the game is over.
These results show that learning features purely from images, without any supervision, can be beneficial for width-based planning. In particular, using the learned representations in RolloutIW(1) leads to generally higher scores in the Atari 2600 domain. 
For reference, in \cref{tab:atari-piiw-rainbow} we provide an additional comparison of our method with other related approaches. However, these significantly differ from VAE-IW in some aspects and are therefore not directly comparable.

Note that the performance of VAE-IW depends on the quality and expressiveness of the features extracted by the VAE, which in turn depend on the images the VAE is trained on. Crucially, since we collect data by running RolloutIW with B-PROST features, the performance of VAE-IW is constrained by the effectiveness of the data collection algorithm. The VAE features will not necessarily be meaningful on parts of the game that are significantly different from those in the training set. Surprisingly, however, our method significantly outperforms the baseline that was used for data collection, providing even stronger evidence that the compact set of features learned by a VAE can be successfully utilized in width-based planning.
This form of bootstrapping can be iterated: images collected by VAE-IW could be used for re-training or fine-tuning VAEs, potentially leading to further performance improvements. Although in principle the first iteration of VAEs could be trained from images collected via random play, this is not a viable solution in hard-exploration problems such as many Atari games \cite{ecoffet2019go}.

In addition, there appears to be a significant negative correlation between the average number of true features and the average number of expanded nodes per planning phase (Spearman rank correlation of $-0.51$, p-value $< 0.001$). In other words, in domains where the VAE extracts on average more true features, the planning algorithm tends to expand fewer nodes. Thus, it could be potentially fruitful to further investigate the interplay between the average number of true features, the meaningfulness and usefulness of the features, and the efficiency of the width-based planning algorithm.

\subsection{Ablation Studies}

The performance of VAE-IW depends on hyperparameters of the planning algorithm and the probabilistic model.  Here we investigate the effect of some of these parameters. Tables with full results are reported in the Appendix.

\paragraph{Modeling choices.}

One of the major modeling choices is the dimensionality of the latent space, and the spatial structure $H \times W \times C$ of the latent variables. Both of these factors are tightly coupled with the neural architecture underlying the inference and generative networks. As there is no clear heuristic, we explored different neural architectures and latent space sizes. Based on the performance on a few selected domains, we chose two representative settings, with latent space size $15 \times 15 \times 20$ and $4 \times 4 \times 200$ (see the Appendix for further details). In Table~\ref{tab:atari-model-comp} we compare the performance of RA VAE-IW(1) on these two configurations, keeping the rest of the hyperparameters fixed as the ones used in Table~\ref{tab:atari-lookahead-0.5}. While overall the $15 \times 15 \times 20$ configuration leads to a higher score in most domains, the effect of this modeling choice seems to significantly depend on the domain.

An alternative to directly training a discrete latent variable model is to train a continuous one and then quantize the inferred representations at planning time. We investigated this possibility by training a continuous VAE with latent space size $15 \times 15 \times 5$ and independent prior $\mathcal{N}(0, 1)$ for each latent variable. When planning, we quantize the inferred posterior means into $2^4$ or $2^6$ bins of equal probability under the prior. This leads to 4 or 6 bits per latent variable, and hence 4,500 and 6,750 boolean features, respectively. In Table~\ref{tab:continuous}, we observe that this approach, although generally competitive, tends to be inferior to training a discrete model directly.

As previously mentioned, we consider the framework of $\beta$-VAEs in which $\beta$ controls the trade-off between the reconstruction accuracy and the amount of information encoded in the latent space. For our purposes, $\beta$ has to be small enough that the model can capture all relevant details in an image. In practice, we decreased $\beta$ until the model generated sufficiently accurate reconstructions on a selection of Atari games. Table~\ref{tab:atari-comparison} reports the performance of VAE-IW(1) when varying $\beta \in \{10^{-4}, 10^{-3}\}$, and shows that the effect of varying $\beta$ depends on the domain. 
Intuitively, while a higher $\beta$ can be detrimental for the reconstructions and thus also for the informativeness of the learned features, it may lead to better representations in less visually challenging domains. In practice, one could for example train VAEs with different regularization strengths, and do unsupervised model selection separately for each domain by using the ``elbow method'' \cite{ketchen1996application} on the reconstruction loss or other relevant metrics.

\paragraph{Planning parameters.}
Regardless of the features used for planning, the performance of VAE-IW depends on the variant of RolloutIW being used, and on its parameters. 
In Table~\ref{tab:atari-ra_no-ra} we compare the average score of VAE-IW(1) with and without risk aversion and observe that the risk-averse variant achieves an equal or higher average score in 68\% of the domains.
Table~\ref{tab:atari-no-ra} shows the results of VAE-IW(1) and RolloutIW(1) with B-PROST features, similarly to Table~\ref{tab:atari-lookahead-0.5}, except that both methods are run \emph{without} risk aversion. With this modification, our method still obtains a higher average score in the majority of games (68\%).

Another crucial planning parameter is the time budget for planning at each time step. While the main results are based on a 0.5s budget, we also consider a 32s budget, following \citet{rolloutiw-orginal}. In Table~\ref{tab:higher_time_budget} we observe that, unsurprisingly, the 32s budget leads to better results in most domains (79\%). 
Interestingly, increasing the planning budget does not seem to affect the average rollout depth, while the average number of expanded nodes in each planning phase grows significantly. This behaviour is consistently observed in all tested domains (see Figures~\ref{fig:mean_expanded_nodes} and ~\ref{fig:mean_depth_nodes} in Appendix) and points to the fact that increasing the planning budget improves results mostly by allowing more rollouts.

In Table~\ref{tab:higher_time_budget_bprost} we compare the average scores obtained by VAE-IW(1) and RolloutIW(1) with B-PROST features, using a 32s planning budget. Once again, using the compact features learned by a VAE seems to be beneficial, as VAE-IW(1) obtains the highest average score in 62\% of the games. However, the margin here is smaller than in the experiments with a budget of 0.5s, where VAE-IW(1) outperformed RolloutIW(1) B-PROST in 73\% of the games.

%% file: tables/human_comp.tex
\resizebox{\columnwidth}{!}{%
\begin{tabular}{lrrrrrrrl}

\toprule
Algorithm & Human & IW &  RolloutIW & VAE-IW \\
 Features & & B-PROST &  B-PROST & VAE \\

\midrule
Alien            & 6,875                & 1,316          & 7,170           & \textbf{7,744}     \\
Amidar           & 1,676                & 48             & 1,049.2           & \textbf{1,380.3}     \\
Assault          & 1,496                & 268.8            & 336             & \textbf{1,291.9}     \\
Asterix          & 8,503                & 1,350          & 46,100          & \textbf{999,500*}   \\
Asteroids        & 13,157               & 840            & 4,698           & \textbf{12,647}    \\
Atlantis         & 29,028               & 33,160         & 122,220         & \textbf{1,977,520*} \\
BankHeist        & 734.4                   & 24             & 242             & \textbf{289}       \\
BattleZone       & 37,800               & 6,800          & 74,600          & \textbf{115,400}   \\
BeamRider        & 5,775                & 715.2            & 2,552.8           & \textbf{3,792}     \\
Berzerk          & \multicolumn{1}{l}{} & 280            & \textbf{1,208}  & 863                \\
Bowling          & 154.8                   & 30.6             & 44.2              & \textbf{54.4}        \\
Boxing           & 4.3                     & \textbf{99.4}    & 99.2              & 89.9                 \\
Breakout         & 31.8                    & 1.6              & \textbf{86.2}     & 45.7                 \\
Centipede        & 11,963               & 88,890         & 56,328          & \textbf{428,451.5*}   \\
ChopperCommand   & 9,882                & 1,760          & \textbf{9,820}  & 4,190              \\
CrazyClimber     & 35,411               & 16,780         & 40,440          & \textbf{901,930*}   \\
DemonAttack      & 3,401                & 106            & 6,958           & \textbf{285,867.5*}   \\
DoubleDunk       & -15.5                   & -22            & 3.2               & \textbf{8.6}         \\
ElevatorAction   & \multicolumn{1}{l}{} & 1,080          & 0               & \textbf{40,000*}    \\
Enduro           & 309.6                   & 2.6              & \textbf{145.8}    & 55.5                 \\
FishingDerby     & 5.5                     & -83.8            & -77             & \textbf{-20}       \\
Freeway          & 29.6                    & 0.6              & 2               & \textbf{5.3}         \\
Frostbite        & 4,335                & 106            & 146             & \textbf{259}       \\
Gopher           & 2,321                & 1,036          & 8,388           & \textbf{8,484}     \\
Gravitar         & 2,672                & 380            & 1,660           & \textbf{1,940}     \\
IceHockey        & 0.9                     & -13.6            & -12.4             & \textbf{37.2}        \\
Jamesbond        & 406.7                   & 40             & \textbf{10,760} & 3,035              \\
Kangaroo         & 3,035                & 160            & \textbf{1,880}  & 1,360              \\
Krull            & 2,395                & 3,206.8          & 2,091.8           & \textbf{3,433.9}     \\
KungFuMaster     & 22,736               & 440            & 2,620           & \textbf{4,550}     \\
MontezumaRevenge & 4,367                & 0              & \textbf{100}    & 0                  \\
MsPacman         & 15,693               & 2,578          & 15,115          & \textbf{17,929.8}    \\
NameThisGame     & 4,076                & 7,070          & 6,558           & \textbf{17,374}    \\
Phoenix          & \multicolumn{1}{l}{} & 1,266          & \textbf{6,790}  & 5,919              \\
Pitfall          & \multicolumn{1}{l}{} & -8.6             & -302.8            & \textbf{-5.6}        \\
Pong             & 9.3                     & -20.8            & -4.2              & \textbf{4.2}         \\
PrivateEye       & 69,571               & \textbf{2,690.8} & -480            & 80                 \\
Qbert            & 13,455               & 515            & \textbf{15,970} & 3,392.5              \\
Riverraid        & 13,513               & 664            & 6,288           & \textbf{6,701}     \\
RoadRunner       & 7,845                & 200            & \textbf{31,140} & 2,980              \\
Robotank         & 11.9                    & 3.2              & \textbf{31.2}     & 25.6                 \\
Seaquest         & 20,182               & 168            & \textbf{2,312}  & 842                \\
Skiing           & \multicolumn{1}{l}{} & -16,511        & -16,006.8         & \textbf{-10,046.9}   \\
Solaris          & \multicolumn{1}{l}{} & 1,356          & 1,704           & \textbf{7,838}     \\
SpaceInvaders    & 1,652                & 280            & 1,149           & \textbf{2,574}     \\
StarGunner       & 10,250               & 840            & \textbf{14,900} & 1,030              \\
Tennis           & -8.9                    & -23.4            & -5.4              & \textbf{4.1}         \\
TimePilot        & 5,925                & 2,360          & 3,540           & \textbf{32,840}    \\
Tutankham        & 167.7                   & 71.2             & 135.6             & \textbf{177}       \\
UpNDown          & 9,082                & 928            & 34,668          & \textbf{762,453*}   \\
Venture          & 1,188                & 0              & \textbf{60}     & 0                  \\
VideoPinball     & 17,298               & 28,706.4         & 216,468.6         & \textbf{373,914.3}   \\
WizardOfWor      & 4,757                & 5,660          & 43,860          & \textbf{199,900}   \\
YarsRevenge      & \multicolumn{1}{l}{} & 6,352.6          & 7,848.8           & \textbf{96,053.3}    \\
Zaxxon           & 9,173                & 0              & 15,500          & \textbf{15,560}    \\
\midrule
\# $>$ Human  & n/a  & 4 & 22 & 25  \\
\# $>$ 75\% Human  & n/a  & 4 & 26 & 28  \\
\midrule
\# best in game & n/a & 2 & 14 &  39 \\
\bottomrule
\end{tabular}%
}
\caption{Score comparison in 55 Atari games. Scores in bold indicate the best width-based method, and a star indicates that the limit on executed actions was reached at least once. 
RolloutIW and VAE-IW are run with risk aversion.
} 
\label{tab:atari-lookahead-0.5}

%% file: sections/related_work.tex
\section{Related Work}

Variational Autoencoders (VAEs) have been extensively used for representation learning \cite{kingma2019introduction} as their amortized inference network lends itself naturally to this task.
In the context of automated planning, \citet{DBLP:journals/corr/AsaiF17} proposed a State Autoencoder (SAE) for propositional symbolic grounding. SAEs are in fact VAEs with Bernoulli latent variables. They are trained by maximizing a modified ELBO with an additional entropy regularizer, defined as twice the negative KL divergence. Thus, the objective function being maximized is the ELBO~\eqref{eq:elbo} with the sign of the KL flipped. 
%
A variation of SAE, the zero-suppressed state autoencoder \cite{asai2019stable}, adds a further regularization term to the propositional representation, leading to more stable features.

\citet{zhang2018composable} learn state features with supervised learning, and then plan in the feature space with a learned transition graph. 
\citet{konidaris2018skills} specify a set of skills for a task, and automatically extract state representations from raw observations. 
\citet{kurutach2018learning} use generative adversarial networks to learn structured representations of images and a deterministic dynamics model, and plan with graph search.
\citet{RLIW} proposed $\pi$-IW, a variant of RolloutIW where a neural network guides the action selection process in the rollouts, which would otherwise be random.
Moreover, $\pi$-IW plans using features obtained from the last hidden layer of the policy network, instead of B-PROST.

Atari games have also been extensively used as a testbed in model-free \citep{dqn, dqn2,a3c,ppo,ga3c,acktr,vtrace, rainbow,kapturowski2018recurrent} and model-based \citep{vpn,dyna_dqn,kaiser2019model,schrittwieser2019mastering,badia2020agent57} deep reinforcement learning. 

%% file: sections/conclusion.tex
\section{Conclusions}
We have introduced a novel combination of width-based planning with learning techniques. The learning employs a VAE to learn relevant features in Atari games given images of screen states as training data. The planning is done with RolloutIW(1) using the features learned by the VAE. 
Our approach reduces the size of the feature set from the 20.5 million B-PROST features used in previous work in connection with RolloutIW, to only 4,500. Our algorithm, VAE-IW, outperforms the previous methods, proving that VAEs can learn meaningful representations that can be effectively used for planning with RolloutIW.

In VAE-IW, the symbolic representations are learned from data collected by RolloutIW using B-PROST features. Increasing the diversity and quality of the training data could potentially lead to better representations, and thus better planning results.
One possible way to achieve this could be to iteratively retrain and fine-tune the VAEs based on data collected by the current iteration of VAE-IW. 
The quality of the representations could also be improved by using more expressive discrete models, for example with a hierarchy of discrete latent variables \cite{van2017neural,razavi2019generating}.
Similarly, in the ablation study where we compare this approach with quantizing continuous representations, more flexible models~\cite{ranganath2016hierarchical,sonderby2016ladder,vahdat2020nvae} and more advanced quantization strategies could be employed.
Finally, we can expect further improvements orthogonal to this work, by learning a rollout policy for more effective action selection as in \citet{RLIW}.

%% file: sections/appendix.tex
\section{Appendix}

\begin{figure*}[ht]
    \centering
    \includegraphics[width=\textwidth]{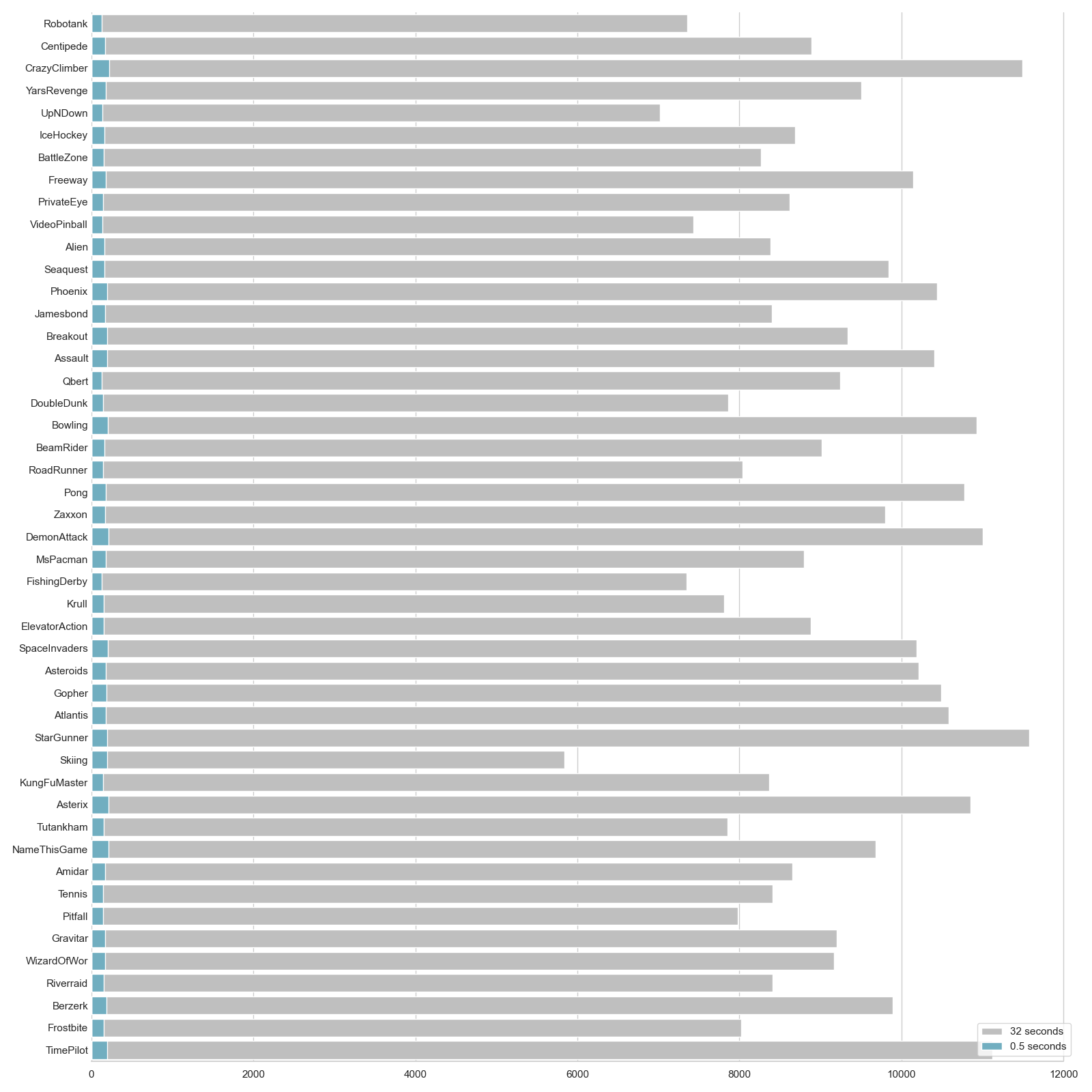}
    \caption{Mean number of expanded nodes during one planning phase for each domain (one run per domain). The comparison is between the 0.5 second RA VAE-IW and 32 second RA VAE-IW.}
    \label{fig:mean_expanded_nodes}
\end{figure*}

\begin{figure*}[ht]
    \centering
    \includegraphics[width=\textwidth]{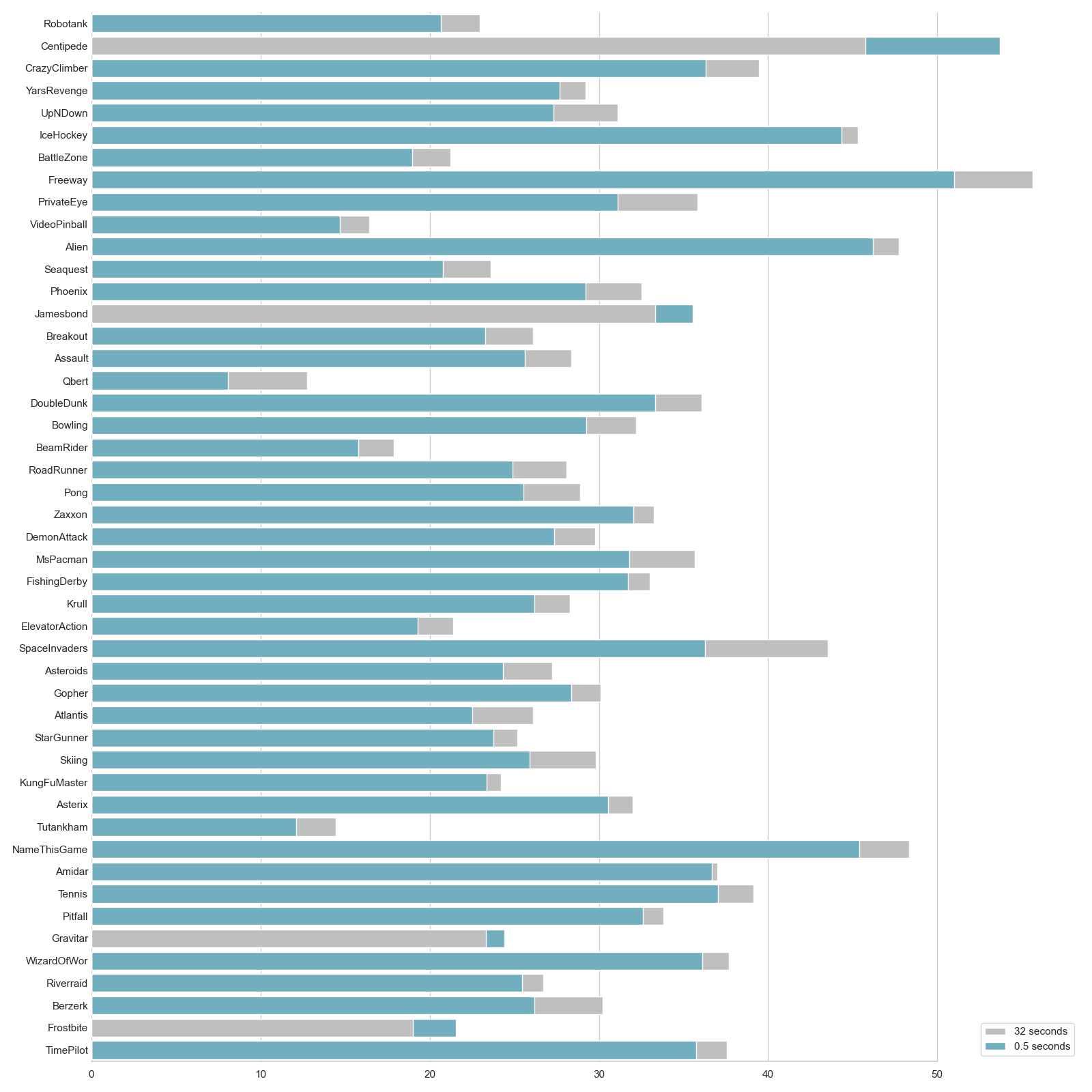}
    \caption{Mean of the maximum rollout length for each domain (one run per domain). The comparison is between the 0.5 second RA VAE-IW and 32 second RA VAE-IW.}
    \label{fig:mean_depth_nodes}
\end{figure*}

In this appendix we provide implementation details and additional results. 
We optimize VAEs with Adam~\cite{kingma2014adam} with default parameters and learning rate $10^{-4}$. The $210 \times 160$ images are grayscaled and downsampled to $128 \times 128$. 
VAE-IW(1) experiments were performed with 4GB of RAM on one core of a Xeon Gold 6126 CPU, and a Tesla V100 GPU.
In \cref{tab:constant} we summarize the default hyperparameters of our main experiments, and in
\crefrange{tab:residualblock}{tab:decoder15} we detail the architectures of the VAEs. We used Bernoulli decoders, i.e. the reconstruction loss is the binary cross-entropy. 
Then, in \crefrange{tab:atari-model-comp}{tab:higher_time_budget_bprost} we report the results of our ablation studies on a subset of 47 games. While the main VAE-IW results (also reported in \cref{tab:atari-lookahead-0.5}) are averages of 10 runs, all other results in this appendix are averages of 5 runs. 
In \cref{tab:atari-piiw-rainbow} we compare our main results from \cref{tab:atari-lookahead-0.5} with $\pi$-IW \cite{RLIW} and Rainbow \cite{rainbow}, which is a DQN-based deep reinforcement learning algorithm. These comparisons are useful to get a sense of how VAE-IW stands against existing approaches for Atari, but should be taken with a grain of salt as these methods (especially Rainbow) differ significantly from VAE-IW.
Finally, Figures~\ref{fig:mean_expanded_nodes} and~\ref{fig:mean_depth_nodes} show additional statistics on the width-based search in terms of number of expanded nodes and their depth.

\vspace{5pt}
\begin{table}[ht]
    \centering
    \begin{tabular}{l l}
        \toprule
        \textbf{Parameter} & \textbf{Value} \\
        \midrule
        Batch size & 64\\
        Learning rate & $10^{-4}$\\
        Latent space size & $15 \times 15 \times 20 = 4500$ \\
        $\tau$ & 0.5\\
        $\beta$ & $10^{-4}$\\
        $\mu$ & $0.5$\\
        $\alpha$ & 50000\\
        $\gamma$ & 0.99 \\
        $\lambda$ & 0.9 \\
        Frame skip & 15\\
        Planning budget & 0.5s\\
        \bottomrule
    \end{tabular}
    \caption{Default hyperparameters for VAE-IW experiments.}
    \label{tab:constant}
\end{table}
\begin{table}[ht]
    \centering 
    \begin{tabular}{l}
        \toprule
        \textbf{Residual block}\\
        \midrule
        BatchNorm\\
        LeakyReLU(0.01)\\
        Conv $3 \times 3$, 64 channels, padding=1\\
        Dropout(0.2)\\
        BatchNorm\\
        LeakyReLU(0.01)\\
        Conv $3 \times 3$, 64 channels, padding=1\\
        Dropout(0.2)\\
        Add to block input \\
        LeakyReLU(0.01)\\
        \bottomrule
    \end{tabular}
    \caption{Architecture of a residual block.}
    \label{tab:residualblock}
\end{table}
\begin{table}[ht]
    \centering 
    \begin{tabular}{l}
        \toprule
        \textbf{Encoder $4 \times 4$}\\
        \midrule
        Conv $3 \times 3$, 64 channels, stride=2\\
        Residual block\\
        Conv $3 \times 3$, 64 channels, stride=2\\
        Residual block\\
        Conv $3 \times 3$, 64 channels, stride=2\\
        Residual block\\
        Conv $3 \times 3$, 64 channels, stride=2\\
        Residual block\\
        Conv $3 \times 3$, $200$ channels, stride=2, padding=1\\
        \bottomrule
    \end{tabular}
    \caption{Encoder with output of spatial size $4\times 4$.}
    \label{tab:encoder4}
\end{table}
\begin{table}[ht]
    \centering 
    \begin{tabular}{l}
        \toprule
        \textbf{Decoder $4 \times 4$}\\
        \midrule
        ConvT $3 \times 3$, 64 channels, stride=2\\
        Residual block\\
        ConvT $3 \times 3$, 64 channels, stride=2\\
        Residual block\\
        ConvT $3 \times 3$, 64 channels, stride=2\\
        Residual block\\
        ConvT $3 \times 3$, 64 channels, stride=2\\
        Residual block\\
        ConvT $3 \times 3$, 1 channel, stride=2\\
        Crop $128 \times 128$ \\
        Sigmoid\\
        \bottomrule
    \end{tabular}
    \caption{Decoder with input of spatial size $4\times 4$. ConvT denotes transposed convolutional layers.}
    \label{tab:decoder4}
\end{table}
\begin{table}[ht]
    \centering 
    \begin{tabular}{l}
        \toprule
        \textbf{Encoder $15 \times 15$}\\
        \midrule
        Conv $4 \times 4$, 64 channels, stride=2\\
        Residual block\\
        Conv $4 \times 4$, 64 channels, stride=2\\
        Residual block\\
        Conv $3 \times 3$, $20$ channels, stride=2, padding=1\\
        \bottomrule
    \end{tabular}
    \caption{Encoder with output of spatial size $15\times 15$.}
    \label{tab:encoder15}
\end{table}
\begin{table}[ht]
    \centering 
    \begin{tabular}{l}
        \toprule
        \textbf{Decoder $15 \times 15$}\\
        \midrule
        ConvT $3 \times 3$, 64 channels, stride=2\\
        Residual block\\
        ConvT $4 \times 4$, 64 channels, stride=2\\
        Residual block\\
        ConvT $4 \times 4$, 1 channel, stride=2\\
        Crop $128 \times 128$ \\
        Sigmoid\\
        \bottomrule
    \end{tabular}
    \caption{Decoder with input of spatial size $15\times 15$. ConvT denotes transposed convolutional layers.}
    \label{tab:decoder15}
\end{table}

\input{tables/model_comp}
\input{tables/continuous}

\input{tables/beta}
\input{tables/ra_and_no_ra}

\input{tables/no_ra}
\input{tables/increase_budget}
\input{tables/pi-iw_rainbow}


%% file: tables/model_comp.tex
\begin{table}[ht]
\begin{center}
\begin{small}
\begin{tabular}{lrrrrrrrl}

\toprule
Algorithm& RA VAE-IW & RA VAE-IW \\
    $\beta$ & $10^{-4}$ & $10^{-4}$\\
Planning horizon&  0.5s & 0.5s \\
Features &   VAE $15 \times 15 \times 20$ &  VAE $4 \times 4 \times 200$ \\
\midrule
Alien           & \textbf{7,744.0}     & 7,592.0              \\
Amidar          & 1,380.3              & \textbf{1,526.6}     \\
Assault         & 1,291.9              & \textbf{1,308.6}     \\
Asterix         & \textbf{999,500.0}   & \textbf{999,500.0}   \\
Asteroids       & \textbf{12,647.0}    & 8,852.0              \\
Atlantis        & \textbf{1,977,520.0*} & 1,912,700.0*          \\
Battle zone     & 115,400.0            & \textbf{228,000.0}   \\
Beam rider      & \textbf{3,792.0}     & 3,450.0              \\
Berzerk         & \textbf{863.0}       & 752.0                \\
Bowling         & \textbf{54.4}        & 47.6                 \\
Breakout        & \textbf{45.7}        & 28.4                 \\
Centipede       & \textbf{428,451.5*}   & 253,823.6*            \\
Crazy climber   & 901,930.0*            & \textbf{930,420.0*}   \\
Demon attack    & 285,867.5*            & \textbf{292,099.0*}   \\
Double dunk     & \textbf{8.6}         & 6.0                  \\
Elevator action & 40,000.0*             & \textbf{109,680.0*}   \\
Fishing derby   & \textbf{-20.0}       & -31.2                \\
Freeway         & \textbf{5.3}         & 2.0                  \\
Frostbite       & \textbf{259.0}       & 244.0                \\
Gopher          & 8,484.0              & \textbf{8,504.0}     \\
Gravitar        & \textbf{1,940.0}     & 1,560.0              \\
Ice hockey      & \textbf{37.2}        & \textbf{37.2}        \\
James bond 007  & 3,035.0              & \textbf{11,000.0}    \\
Krull           & \textbf{3,433.9}     & 3,219.0              \\
Kung-fu master  & \textbf{4,550.0}     & 3,600.0              \\
Ms. Pac-man     & \textbf{17,929.8}    & 15,066.8             \\
Name this game  & \textbf{17,374.0}    & 13,670.0             \\
Phoenix         & 5,919.0              & \textbf{9,210.0}     \\
Pitfall!        & \textbf{-5.6}        & -9.6                 \\
Pong            & \textbf{4.2}         & 3.6                  \\
Private eye     & 80.0                 & \textbf{115.4}       \\
Q*bert          & 3,392.5              & \textbf{4,935.0}     \\
River raid      & 6,701.0              & \textbf{6,790.0}     \\
Road Runner     & \textbf{2,980.0}     & 2,320.0              \\
Robotank        & 25.6                 & \textbf{45.2}        \\
Seaquest        & 842.0                & \textbf{1,084.0}     \\
Skiing          & \textbf{-10,046.9}   & -11,906.4            \\
Space invaders  & 2,574.0              & \textbf{2,753.0}     \\
Stargunner      & 1,030.0              & \textbf{1,200.0}     \\
Tennis          & \textbf{4.1}         & -6.6                 \\
Time pilot      & \textbf{32,840.0}    & 23,460.0             \\
Tutankham       & \textbf{177.0}       & 158.4                \\
Up'n down       & \textbf{762,453.0*}   & 627,706.0*            \\
Video pinball   & \textbf{373,914.3}   & 248,101.2            \\
Wizard of wor   & \textbf{199,900.0*}   & 111,580.0            \\
Yars' revenge   & 96,053.3             & \textbf{97,004.6}    \\
Zaxxon          & 15,560.0             & \textbf{17,360.0}    \\
\midrule

\# best & 29 & 20  \\

\bottomrule

\end{tabular}
\end{small}
\end{center}
\caption{Score comparison of VAE-IW with latent space size $15\times15\times20$ and $4\times4\times 200$. Scores in bold indicate the best method, and scores with a star indicate that the algorithm reached the limit on executed actions at least once. Ties are counted as won for both methods.} 
\label{tab:atari-model-comp}
\end{table}

%% file: tables/continuous.tex
\begin{table}
\centering

\resizebox{\columnwidth}{!}{%
\begin{tabular}{lrrrrrrrl}

\toprule
Algorithm & RA VAE-IW & RA VAE-IW & RA VAE-IW \\
Model type & Continuous & Continuous & Discrete \\
Quantization & 6 bits & 4 bits & --- \\

\midrule
Alien          & 6,988.0              & 7,532.0              & \textbf{7,744.0}     \\
Amidar         & 784.0                & 612.2                & \textbf{1,380.3}     \\
Assault        & 1,286.0              & \textbf{1,357.4}     & 1,291.9              \\
Asterix        & \textbf{999,500.0*}   & \textbf{999,500.0*}   & \textbf{999,500.0*}   \\
Asteroids      & 2,886.0              & 2,670.0              & \textbf{12,647.0}    \\
Atlantis       & \textbf{1,999,040.0*} & 1,997,760.0 *         & 1,977,520.0*          \\
BattleZone     & 90,800.0             & 83,400.0             & \textbf{115,400.0}   \\
BeamRider      & \textbf{4,380.0}     & 3,810.0              & 3,792.0              \\
Berzerk        & \textbf{1,032.0}     & \textbf{1,032.0}     & 863.0                \\
Bowling        & 48.4                 & 43.4                 & \textbf{54.4}        \\
Breakout       & 73.2                 & \textbf{158.4}       & 45.7                 \\
Centipede      & 1,191,414.0*          & \textbf{1,214,556.4*} & 428,451.5*            \\
CrazyClimber   & 225,980.0*            & 264,160.0*            & \textbf{901,930.0*}   \\
DemonAttack    & 291,138.0*            & \textbf{299,555.0*}   & 285,867.5*            \\
DoubleDunk     & 3.4                  & 0.2                  & \textbf{8.6}         \\
ElevatorAction & 35,580.0             & 38,000.0             & \textbf{40,000.0}    \\
FishingDerby   & -26.2                & -25.6                & \textbf{-20.0}       \\
Freeway        & 7.8                  & \textbf{8.2}         & 5.3                  \\
Frostbite      & \textbf{282.0}       & 238.0                & 259.0                \\
Gopher         & 6,772.0              & 7,580.0              & \textbf{8,484.0}     \\
Gravitar       & 1,840.0              & \textbf{3,380.0}     & 1,940.0              \\
IceHockey      & 29.0                 & 34.8                 & \textbf{37.2}        \\
Jamesbond      & 780.0                & 1,500.0              & \textbf{3,035.0}     \\
Krull          & 3,175.8              & \textbf{3,625.2}     & 3,433.9              \\
KungFuMaster   & \textbf{6,740.0}     & \textbf{6,740.0}     & 4,550.0              \\
MsPacman       & \textbf{24,555.0}    & 17,630.8             & 17,929.8             \\
NameThisGame   & \textbf{17,594.0}    & 16,046.0             & 17,374.0             \\
Phoenix        & 5,532.0              & 5,902.0              & \textbf{5,919.0}     \\
Pitfall        & -9.6                 & -47.8                & \textbf{-5.6}        \\
Pong           & 1.8                  & \textbf{4.6}         & 4.2                  \\
PrivateEye     & -102.6               & -107.8               & \textbf{80.0}        \\
Qbert          & 2,795.0              & \textbf{5,575.0}     & 3,392.5              \\
Riverraid      & \textbf{7,838.0}     & 7,382.0              & 6,701.0              \\
RoadRunner     & 2,600.0              & \textbf{6,900.0}     & 2,980.0              \\
Robotank       & \textbf{34.8}        & 23.6                 & 25.6                 \\
Seaquest       & \textbf{3,852.0}     & 2,016.0              & 842.0                \\
Skiing         & -28,186.0            & -29,598.6            & \textbf{-10,046.9}   \\
SpaceInvaders  & 2,058.0              & 1,862.0              & \textbf{2,574.0}     \\
StarGunner     & 1,180.0              & \textbf{1,400.0}     & 1,030.0              \\
Tennis         & -1.8                 & 0.4                  & \textbf{4.1}         \\
TimePilot      & 21,300.0             & 23,500.0             & \textbf{32,840.0}    \\
Tutankham      & 162.4                & 157.2                & \textbf{177.0}       \\
UpNDown        & 850,624.0*            & \textbf{862,708.0*}   & 762,453.0*            \\
VideoPinball   & 370,304.6            & \textbf{477,225.2}   & 373,914.3            \\
WizardOfWor    & 109,140.0            & 151,060.0            & \textbf{199,900.0*}   \\
YarsRevenge    & 117,236.2            & \textbf{137,892.4}   & 96,053.3             \\
Zaxxon         & 15,260.0             & \textbf{15,860.0}    & 15,560.0             \\
\midrule

\# best in game & 10 & 18 & 23 \\
\bottomrule
\end{tabular}%
}
\caption{Score comparison of RA VAE-IW with discrete (default model with latent space of size $15 \times 15 \times 20 = 4500$) and continuous random variables. In the continuous case, the latent space has size $15 \times 15 \times 5$, and then each random variable is quantized in $2^4$ or $2^6$ bins with equal probability under the prior. This leads to 4 or 6 bits per random variable, and hence 4,500 and 6,750 binary features, respectively. All other parameters are the default ones from the main text.
} 
\label{tab:continuous}
\end{table}

%% file: tables/beta.tex
\begin{table}[ht]
\begin{center}
\begin{small}
\begin{tabular}{lrrrl}

\toprule
Algorithm  & RA VAE-IW & RA VAE-IW \\
$\beta$ & $10^{-3}$ & $10^{-4}$\\
Planning horizon & 0.5s & 0.5s  \\
Features &  VAE $15 \times 15 \times 20$ &  VAE $15 \times 15 \times 20$ \\
\midrule

Alien           & 4,902.0              & \textbf{7,744.0}     \\
Amidar          & 1,186.2              & \textbf{1,380.3}     \\
Assault         & \textbf{1,468.2}     & 1,291.9              \\
Asterix         & \textbf{999,500.0*}   & \textbf{999,500.0*}   \\
Asteroids       & 7,204.0              & \textbf{12,647.0}    \\
Atlantis        & \textbf{1,978,540.0*} & 1,977,520.0*          \\
Battle zone     & \textbf{406,600.0}   & 115,400.0            \\
Beam rider      & \textbf{4,377.6}     & 3,792.0              \\
Berzerk         & 774.0                & \textbf{863.0}       \\
Bowling         & 35.0                 & \textbf{54.4}        \\
Breakout        & \textbf{53.4}        & 45.7                 \\
Centipede       & 296,791.4*           & \textbf{428,451.5*}   \\
Crazy climber   & \textbf{976,580.0*}   & 901,930.0*            \\
Demon attack    & \textbf{301,886.0*}   & 285,867.5*            \\
Double dunk     & 7.4                  & \textbf{8.6}         \\
Elevator action & \textbf{68,920.0*}    & 40,000.0*             \\
Fishing derby   & \textbf{-15.6}       & -20.0                \\
Freeway         & 4.0                  & \textbf{5.3}         \\
Frostbite       & \textbf{262.0}       & 259.0                \\
Gopher          & 5,420.0              & \textbf{8,484.0}     \\
Gravitar        & \textbf{2,300.0}     & 1,940.0              \\
Ice hockey      & 34.0                 & \textbf{37.2}        \\
James bond 007  & 630.0                & \textbf{3,035.0}     \\
Krull           & \textbf{3,486.4}     & 3,433.9              \\
Kung-fu master  & \textbf{4,960.0}     & 4,550.0              \\
Ms. Pac-man     & 17,483.0             & \textbf{17,929.8}    \\
Name this game  & 15,120.0             & \textbf{17,374.0}    \\
Phoenix         & 5,524.0              & 5,919.0              \\
Pitfall!        & -6.8                 & \textbf{-5.6}        \\
Pong            & -2.2                 & \textbf{4.2}         \\
Private eye     & 40.0                 & \textbf{80.0}        \\
Q*bert          & \textbf{8,040.0}     & 3,392.5              \\
River raid      & 6,078.0              & \textbf{6,701.0}     \\
Road Runner     & 2,080.0              & \textbf{2,980.0}     \\
Robotank        & \textbf{44.6}        & 25.6                 \\
Seaquest        & 316.0                & \textbf{842.0}       \\
Skiing          & -11,027.6            & \textbf{-10,046.9}   \\
Space invaders  & \textbf{2,721.0}     & 2,574.0              \\
Stargunner      & \textbf{1,100.0}     & 1,030.0              \\
Tennis          & -16.6                & \textbf{4.1}         \\
Time pilot      & 30,920.0             & \textbf{32,840.0}    \\
Tutankham       & 165.8                & \textbf{177.0}       \\
Up'n down       & 682,080.0*            & \textbf{762,453.0*}   \\
Video pinball   & \textbf{445,085.8}   & 373,914.3            \\
Wizard of wor   & 184,260.0            & \textbf{199,900.0*}   \\
Yars' revenge   & 77,950.8             & \textbf{96,053.3}    \\
Zaxxon          & 10,520.0             & \textbf{15,560.0}    \\

\midrule

\# best &    19                   & 29                      \\

\bottomrule

\end{tabular}
\end{small}
\end{center}
\caption{Score comparison of RA VAE-IW with different values of $\beta$. Scores in bold indicate the best method, and scores with a star indicate that the algorithm reached the limit on executed actions at least once. Ties are counted as won for both methods.} 
\label{tab:atari-comparison}
\end{table}

%% file: tables/ra_and_no_ra.tex
\begin{table}[ht]
\begin{center}
\begin{small}
\begin{tabular}{lrrrrrrrl}
\toprule
Algorithm& VAE-IW & RA VAE-IW \\
$\beta$ & $10^{-4}$ & $10^{-4}$\\
Planning horizon& 0.5s & 0.5s  \\
Features & VAE $15 \times 15 \times 20$ & VAE $15 \times 15 \times 20$\\

\midrule

Alien           & 5,576.0              & \textbf{7,744.0}     \\
Amidar          & 1,093.2              & \textbf{1,380.3}     \\
Assault         & 826.4                & \textbf{1,291.9}     \\
Asterix         & \textbf{999,500.0*}   & \textbf{999,500.0*}   \\
Asteroids       & 772.0                & \textbf{12,647.0}    \\
Atlantis        & 40,620.0             & \textbf{1,977,520.0*} \\
Battle zone     & 91,200.0             & \textbf{115,400.0}   \\
Beam rider      & 3,179.6              & \textbf{3,792.0}     \\
Berzerk         & 566.0                & \textbf{863.0}       \\
Bowling         & \textbf{64.6}        & 54.4                 \\
Breakout        & 13.0                 & \textbf{45.7}        \\
Centipede       & 22,020.2             & \textbf{428,451.5*}   \\
Crazy climber   & 661,460.0            & \textbf{901,930.0*}   \\
Demon attack    & 9,656.0              & \textbf{285,867.5*}   \\
Double dunk     & 7.2                  & \textbf{8.6}         \\
Elevator action & \textbf{76,160.0*}    & 40,000.0*             \\
Fishing derby   & -20.2                & \textbf{-20.0}       \\
Freeway         & \textbf{6.2}         & 5.3                  \\
Frostbite       & 248.0                & \textbf{259.0}       \\
Gopher          & 6,084.0              & \textbf{8,484.0}     \\
Gravitar        & \textbf{2,770.0}     & 1,940.0              \\
Ice hockey      & 36.0                 & \textbf{37.2}        \\
James bond 007  & 640.0                & \textbf{3,035.0}     \\
Krull           & \textbf{3,543.2}     & 3,433.9              \\
Kung-fu master  & \textbf{5,160.0}     & 4,550.0              \\
Ms. Pac-man     & \textbf{20,161.0}    & 17,929.8             \\
Name this game  & 13,332.0             & \textbf{17,374.0}    \\
Phoenix         & 5,328.0              & \textbf{5,919.0}     \\
Pitfall!        & \textbf{0.0}         & -5.6                 \\
Pong            & \textbf{9.2}         & 4.2                  \\
Private eye     & \textbf{139.0}       & 80.0                 \\
Q*bert          & 1,890.0              & \textbf{3,392.5}     \\
River raid      & 4,884.0              & \textbf{6,701.0}     \\
Road Runner     & \textbf{4,720.0}     & 2,980.0              \\
Robotank        & 21.4                 & \textbf{25.6}        \\
Seaquest        & 596.0                & \textbf{842.0}       \\
Skiing          & \textbf{-9,664.8}    & -10,046.9            \\
Space invaders  & 1,962.0              & \textbf{2,574.0}     \\
Stargunner      & \textbf{1,120.0}     & 1,030.0              \\
Tennis          & \textbf{5.2}         & 4.1                  \\
Time pilot      & 24,840.0             & \textbf{32,840.0}    \\
Tutankham       & 167.4                & \textbf{177.0}       \\
Up'n down       & 35,964.0             & \textbf{762,453.0*}   \\
Video pinball   & \textbf{462,619.4}   & 373,914.3            \\
Wizard of wor   & 89,380.0             & \textbf{199,900.0*}   \\
Yars' revenge   & 85,800.6             & \textbf{96,053.3}    \\
Zaxxon          & 7,320.0              & \textbf{15,560.0}    \\

\midrule

\# best & 16 &  32 \\

\bottomrule

\end{tabular}
\end{small}
\end{center}
\caption{Score  comparison  of  VAE-IW  with  and  without risk aversion. Scores in bold indicate the best method, and scores with a star indicate that the algorithm reached the limit on executed actions at least once. Ties are counted as won for both methods. } 
\label{tab:atari-ra_no-ra}
\end{table}

%% file: tables/no_ra.tex
\begin{table}[ht]
\begin{center}
\begin{small}
\begin{tabular}{lrrl}
\toprule
Algorithm& Rollout IW & VAE-IW \\
$\beta$ &  & $10^{-4}$\\
Planning horizon& 0.5s & 0.5s  \\
Features & B-PROST & VAE $15 \times 15 \times 20$\\

\midrule

Alien           & 4,238.0           & \textbf{5,576.0}   \\
Amidar          & 659.8             & \textbf{1,093.2}   \\
Assault         & 285.0             & \textbf{826.4}     \\
Asterix         & 45,780.0          & \textbf{999,500.0} \\
Asteroids       & \textbf{4,344.0}  & 772.0              \\
Atlantis        & \textbf{64,200.0} & 40,620.0           \\
Battle zone     & 39,600.0          & \textbf{91,200.0}  \\
Beam rider      & 2,188.0           & \textbf{3,179.6}   \\
Berzerk         & \textbf{644.0}    & 566.0              \\
Bowling         & 47.6              & \textbf{64.6}      \\
Breakout        & \textbf{82.4}     & 13.0               \\
Centipede       & \textbf{36,980.2} & 22,020.2           \\
Crazy climber   & 39,220.0          & \textbf{661,460.0*} \\
Demon attack    & 2,780.0           & \textbf{9,656.0}   \\
Double dunk     & 3.6               & \textbf{7.2}       \\
Elevator action & 0.0               & \textbf{76,160.0*}  \\
Fishing derby   & -68.0             & \textbf{-20.2}     \\
Freeway         & 2.8               & \textbf{6.2}       \\
Frostbite       & 220.0             & \textbf{248.0}     \\
Gopher          & \textbf{7,216.0}  & 6,084.0            \\
Gravitar        & 1,630.0           & \textbf{2,770.0}   \\
Ice hockey      & -6.0              & \textbf{36.0}      \\
James bond 007  & 450.0             & \textbf{640.0}     \\
Krull           & 1,892.8           & \textbf{3,543.2}   \\
Kung-fu master  & 2,080.0           & \textbf{5,160.0}   \\
Ms. Pac-man     & 9,178.4           & \textbf{20,161.0}  \\
Name this game  & 6,226.0           & \textbf{13,332.0}  \\
Phoenix         & \textbf{5,750.0}  & 5,328.0            \\
Pitfall!        & -81.4             & \textbf{0.0}       \\
Pong            & -7.4              & \textbf{9.2}       \\
Private eye     & -322.0            & \textbf{139.0}     \\
Q*bert          & \textbf{3,375.0}  & 1,890.0            \\
River raid      & \textbf{6,088.0}  & 4,884.0            \\
Road Runner     & 2,360.0           & \textbf{4,720.0}   \\
Robotank        & \textbf{31.0}     & 21.4               \\
Seaquest        & \textbf{980.0}    & 596.0              \\
Skiing          & -15,738.8         & \textbf{-9,664.8}  \\
Space invaders  & \textbf{2,628.0}  & 1,962.0            \\
Stargunner      & \textbf{13,360.0} & 1,120.0            \\
Tennis          & -18.6             & \textbf{5.2}       \\
Time pilot      & 7,640.0           & \textbf{24,840.0}  \\
Tutankham       & 128.4             & \textbf{167.4}     \\
Up'n down       & \textbf{36,236.0} & 35,964.0           \\
Video pinball   & 203,765.4         & \textbf{462,619.4} \\
Wizard of wor   & 37,220.0          & \textbf{89,380.0}  \\
Yars' revenge   & 5,225.4           & \textbf{85,800.6}  \\
Zaxxon          & \textbf{9,280.0}  & 7,320.0           \\

\midrule

\# best & 15 &  32 \\

\bottomrule

\end{tabular}
\end{small}
\end{center}
\caption{Score comparison between RolloutIW B-PROST and VAE-IW, both without risk aversion. Scores in bold indicate the best method, and scores with a star indicate that the algorithm reached the limit on executed actions at least once. Ties are counted as won for both methods.} 
\label{tab:atari-no-ra}
\end{table}

%% file: tables/increase_budget.tex
\begin{table}[ht]
    \centering
\begin{center}
\begin{small}
\begin{tabular}{lrr}
\toprule
Algorithm & VAE-IW & VAE-IW \\
$\beta$ & $10^{-4}$ & $10^{-4}$\\
Planning horizon & 0.5s & 32s \\
Features &  VAE $15 \times 15 \times 20$ &  VAE $15 \times 15 \times 20$ \\
\midrule

Alien          & 5,576.0            & \textbf{8,536.0}   \\
Amidar         & 1,093.2            & \textbf{1,955.8}   \\
Assault        & 826.4              & \textbf{1,338.4}   \\
Asterix        & \textbf{999,500.0} & \textbf{999,500.0}         \\
Asteroids      & 772.0              & \textbf{1,158.0}   \\
Atlantis       & 40,620.0           & \textbf{49,500.0}  \\
BattleZone     & 91,200.0           & \textbf{234,200.0} \\
BeamRider      & 3,179.6            & \textbf{5,580.0}   \\
Berzerk        & \textbf{566.0}     & 554.0              \\
Bowling        & \textbf{64.6}      & 46.8               \\
Breakout       & 13.0               & \textbf{72.4}      \\
Centipede      & 22,020.2           & \textbf{166,244.8} \\
CrazyClimber   & 661,460.0          & \textbf{731,867.2}  \\
DemonAttack    & \textbf{9,656.0}   & 5,397.0            \\
DoubleDunk     & \textbf{7.2}       & 5.2                \\
ElevatorAction & 76,160.0           & \textbf{88,100.0}     \\
FishingDerby   & -20.2              & \textbf{20.6}      \\
Freeway        & 6.2                & \textbf{29.4}      \\
Frostbite      & 248.0              & \textbf{280.0}     \\
Gopher         & 6,084.0            & \textbf{17,604.0}  \\
Gravitar       & \textbf{2,770.0}   & 2,640.0            \\
IceHockey      & 36.0               & \textbf{44.2}      \\
Jamesbond      & 640.0              & \textbf{650.0}     \\
Krull          & 3,543.2            & \textbf{6,664.0}   \\
KungFuMaster   & 5,160.0            & \textbf{20,960.0}  \\
MsPacman       & 20,161.0           & \textbf{25,759.0}  \\
NameThisGame   & 13,332.0           & \textbf{15,276.0}  \\
Phoenix        & 5,328.0            & \textbf{5,960.0}   \\
Pitfall        & \textbf{0.0}       & -0.4               \\
Pong           & 9.2                & \textbf{12.0}      \\
PrivateEye     & 139.0              & \textbf{157.8}     \\
Qbert          & 1,890.0            & \textbf{4,760.0}   \\
Riverraid      & 4,884.0            & \textbf{5,372.0}   \\
RoadRunner     & 4,720.0            & \textbf{8,540.0}   \\
Robotank       & 21.4               & \textbf{24.2}      \\
Seaquest       & \textbf{596.0}     & 324.0              \\
Skiing         & \textbf{-9,664.8}  & -9,705.0           \\
SpaceInvaders  & 1,962.0            & \textbf{2,972.0}   \\
StarGunner     & {1,120.0}          & \textbf{1,180.0}   \\
Tennis         & {5.2}              & \textbf{12.6}      \\
TimePilot      & \textbf{24,840.0}  & 24,220.0           \\
Tutankham      & 167.4              & \textbf{197.2}     \\
UpNDown        & 35,964.0           & \textbf{91,592.0}  \\
VideoPinball   & {462,619.4}        & \textbf{833,518.4} \\
WizardOfWor    & \textbf{89,380.0}  & 76,460.0           \\
YarsRevenge    & 85,800.6           & \textbf{188,551.2} \\
Zaxxon         & 7,320.0            & \textbf{30,200.0} \\

\midrule
\# best & 11 & 37 \\
\bottomrule

\end{tabular}
\caption{Score comparison of VAE-IW with planning budget of 0.5 or 32 seconds. Scores in bold indicate the best method, and scores with a star indicate that the algorithm reached the limit on executed actions at least once. Ties are counted as won for both methods.}
\label{tab:higher_time_budget}

\end{small}
\end{center}
\end{table}

\begin{table}[ht]
    \centering
\begin{center}
\begin{small}
\begin{tabular}{lrr}
\toprule
Algorithm & RolloutIW & VAE-IW \\
$\beta$ &  & $10^{-4}$\\
Planning horizon & 32s & 32s \\
Features & B-PROST &  VAE $15 \times 15 \times 20$ \\
\midrule

Alien          & 6,896.0            & \textbf{8,536.0}   \\
Amidar         & 1,698.6            & \textbf{1,955.8}   \\
Assault        & 319.2              & \textbf{1,338.4}   \\
Asterix        & 66,100.0           & \textbf{417,100.0} \\
Asteroids      & \textbf{7,258.0}   & 1,158.0            \\
Atlantis       & \textbf{151,120.0} & 49,500.0           \\
BattleZone     & \textbf{414,000.0} & 234,200.0          \\
BeamRider      & 2,464.8            & \textbf{5,580.0}   \\
Berzerk        & \textbf{862.0}     & 554.0              \\
Bowling        & 45.8               & \textbf{46.8}      \\
Breakout       & 36.0               & \textbf{72.4}      \\
Centipede      & 65,162.6           & \textbf{166,244.8} \\
CrazyClimber   & 43,960.0           & \textbf{129,840.0} \\
DemonAttack    & \textbf{9,996.0}   & 5,397.0            \\
DoubleDunk     & \textbf{20.0}      & 5.2                \\
ElevatorAction & \textbf{0.0}       & \textbf{0.0}       \\
FishingDerby   & -16.2              & \textbf{20.6}      \\
Freeway        & 12.6               & \textbf{29.4}      \\
Frostbite      & \textbf{5,484.0}   & 280.0              \\
Gopher         & 13,176.0           & \textbf{17,604.0}  \\
Gravitar       & \textbf{3,700.0}   & 2,640.0            \\
IceHockey      & 6.6                & \textbf{44.2}      \\
Jamesbond      & \textbf{22,250.0}  & 650.0              \\
Krull          & 1,151.2            & \textbf{6,664.0}   \\
KungFuMaster   & 14,920.0           & \textbf{20,960.0}  \\
MsPacman       & 19,667.0           & \textbf{25,759.0}  \\
NameThisGame   & 5,980.0            & \textbf{15,276.0}  \\
Phoenix        & \textbf{7,636.0}   & 5,960.0            \\
Pitfall        & -130.8             & \textbf{-0.4}      \\
Pong           & \textbf{17.6}      & 12.0               \\
PrivateEye     & \textbf{3,157.2}   & 157.8              \\
Qbert          & \textbf{8,390.0}   & 4,760.0            \\
Riverraid      & \textbf{8,156.0}   & 5,372.0            \\
RoadRunner     & \textbf{37,080.0}  & 8,540.0            \\
Robotank       & \textbf{52.6}      & 24.2               \\
Seaquest       & \textbf{10,932.0}  & 324.0              \\
Skiing         & -16,477.0          & \textbf{-9,705.0}  \\
SpaceInvaders  & 1,980.0            & \textbf{2,972.0}   \\
StarGunner     & \textbf{15,640.0}  & 1,180.0            \\
Tennis         & -2.2               & \textbf{12.6}      \\
TimePilot      & 8,140.0            & \textbf{24,220.0}  \\
Tutankham      & 184.0              & \textbf{197.2}     \\
UpNDown        & 44,306.0           & \textbf{91,592.0}  \\
VideoPinball   & 382,294.8          & \textbf{833,518.4} \\
WizardOfWor    & 73,820.0           & \textbf{76,460.0}  \\
YarsRevenge    & 9,866.4            & \textbf{188,551.2} \\
Zaxxon         & 22,880.0           & \textbf{30,200.0} \\

\midrule
\# best & 19 & 29 \\
\bottomrule

\end{tabular}
\caption{Score comparison between RolloutIW B-PROST and VAE-IW with planning budget of 32 seconds. Scores in bold indicate the best method, and scores with a star indicate that the algorithm reached the limit on executed actions at least once. Ties are counted as won for both methods.}
\label{tab:higher_time_budget_bprost}

\end{small}
\end{center}
\end{table}

%% file: tables/pi-iw_rainbow.tex
\begin{table}[ht]
\begin{center}
\begin{small}
\begin{tabular}{lrrrrrrrl}
\toprule
Algorithm & Rainbow & $\pi$-IW & RA VAE-IW \\

\midrule

Alien            & 6,022.9                 & 5,081.4                 & \textbf{7,744.0}     \\
Amidar           & 202.8                   & 1,163.6                 & \textbf{1,380.3}     \\
Assault          & \textbf{14,491.7}       & 3,879.3                 & 1,291.9              \\
Asterix          & 280,114.0               & 6,852.0                 & \textbf{999,500.0}   \\
Asteroids        & 2,249.4                 & 2,708.7                 & \textbf{12,647.0}    \\
Atlantis         & 814,684.0               & 140,336.0               & \textbf{1,977,520.0} \\
BankHeist        & \textbf{826.0}          & 324.2                   & 289.0                \\
BattleZone       & 52,040.0                & \textbf{137,500.0}      & 115,400.0            \\
BeamRider        & \textbf{21,768.5}       & 3,025.6                 & 3,792.0              \\
Berzerk          & \textbf{1,793.4}        & 757.2                   & 863.0                \\
Bowling          & 39.4                    & 32.3                    & \textbf{54.4}        \\
Boxing           & 54.9                    & 89.5                    & \textbf{89.9}        \\
Breakout         & \textbf{379.5}          & 175.1                   & 45.7                 \\
Centipede        & 7,160.9                 & 32,531.7                & \textbf{428,451.5}   \\
ChopperCommand   & \textbf{10,916.0}       & 10,538.0                & 4,190.0              \\
CrazyClimber     & 143,962.0               & 101,246.0               & \textbf{901,930.0}   \\
DemonAttack      & 109,670.7               & 8,690.1                 & \textbf{285,867.5}   \\
DoubleDunk       & -0.6                    & \textbf{20.1}           & 8.6                  \\
Enduro           & \textbf{2,061.1}        & 225.5                   & 55.5                 \\
FishingDerby     & \textbf{22.6}           & 18.7                    & -20.0                \\
Freeway          & 29.1                    & \textbf{29.7}           & 5.3                  \\
Frostbite        & \textbf{4,141.1}        & 3,995.6                 & 259.0                \\
Gopher           & 72,595.7                & \textbf{197,496.8}      & 8,484.0              \\
Gravitar         & 567.5                   & \textbf{2,276.0}        & 1,940.0              \\
IceHockey        & -0.7                    & \textbf{47.0}           & 37.2                 \\
Kangaroo         & \textbf{10,841.0}       & 2,216.0                 & 1,360.0              \\
Krull            & \textbf{6,715.5}        & 4,022.5                 & 3,433.9              \\
KungFuMaster     & \textbf{28,999.8}       & 17,406.0                & 4,550.0              \\
MontezumaRevenge & \textbf{154.0}          & 0.0                     & 0.0                  \\
MsPacman         & 2,570.2                 & 9,006.5                 & \textbf{17,929.8}    \\
NameThisGame     & 11,686.5                & 8,738.6                 & \textbf{17,374.0}    \\
Phoenix          & \textbf{103,061.6}      & 5,769.0                 & 5,919.0              \\
Pitfall          & -37.6                   & -85.1                   & \textbf{-5.6}        \\
Pong             & 19.0                    & \textbf{20.9}           & 4.2                  \\
PrivateEye       & 1,704.4                 & \textbf{4,335.7}        & 80.0                 \\
Qbert            & 18,397.6                & \textbf{248,572.5}      & 3,392.5              \\
RoadRunner       & 54,261.0                & \textbf{100,882.0}      & 2,980.0              \\
Robotank         & 55.2                    & \textbf{60.3}           & 25.6                 \\
Seaquest         & \textbf{19,176.0}       & 1,350.4                 & 842.0                \\
Skiing           & -11,685.8               & -26,081.1               & \textbf{-10,046.9}   \\
Solaris          & 2,860.7                 & 4,442.4                 & \textbf{7,838.0}     \\
SpaceInvaders    & \textbf{12,629.0}       & 2,385.9                 & 2,574.0              \\
StarGunner       & \textbf{123,853.0}      & 7,408.0                 & 1,030.0              \\
Tennis           & -2.2                    & -12.2                   & \textbf{4.1}         \\
TimePilot        & 11,190.5                & 10,770.0                & \textbf{32,840.0}    \\
Tutankham        & 126.9                   & \textbf{197.7}          & 177.0                \\
Venture          & \textbf{45.0}           & 0.0                     & 0.0                  \\
VideoPinball     & \textbf{506,817.2}      & 133,521.9               & 373,914.3            \\
WizardOfWor      & 14,631.5                & 44,508.0                & \textbf{199,900.0}   \\
YarsRevenge      & 93,007.9                & 44,864.6                & \textbf{96,053.3}    \\
Zaxxon           & \textbf{19,658.0}       & 12,828.0                & 15,560.0             \\

\midrule

\# best & 20 & 12 & 19 \\

\bottomrule

\end{tabular}
\end{small}
\end{center}
\caption{Score comparisons between RA VAE-IW, $\pi$-IW \cite{RLIW} and Rainbow \cite{rainbow2017}. Scores in bold indicate the best method. Here we only report results for domains where data is available for all 3 methods.} 
\label{tab:atari-piiw-rainbow}
\end{table}